\documentclass[10pt,twocolumn,letterpaper]{article}

\usepackage[pagenumbers]{cvpr} 
\usepackage{graphicx}
\usepackage{amsmath}
\usepackage{amssymb}
\usepackage{booktabs}
\usepackage{multirow}

\usepackage{xcolor}

\usepackage{enumitem}

\usepackage[pagebackref,breaklinks,colorlinks]{hyperref}

\usepackage[capitalize]{cleveref}
\crefname{section}{Sec.}{Secs.}
\Crefname{section}{Section}{Sections}
\Crefname{table}{Table}{Tables}
\crefname{table}{Tab.}{Tabs.}

\def\cvprPaperID{10540} 
\def\confName{CVPR}
\def\confYear{2023}

\begin{document}
%
\title{Scale-Aware Crowd Counting Using a Joint Likelihood Density Map and Synthetic Fusion Pyramid Network}
\author{Yi-Kuan Hsieh\textsuperscript{\rm 1},
    Jun-Wei Hsieh\textsuperscript{\rm 2},
    Yu-Chee Tseng\textsuperscript{\rm 3},
    Ming-Ching Chang\textsuperscript{\rm 4},
    Bor-Shiun Wang\textsuperscript{\rm 5}\\
    \textsuperscript{\rm 1235}National Yang Ming Chiao Tung University,\textsuperscript{\rm 4}University at Albany, State University of New York\\
    \textsuperscript{\rm 1}khjhsnaughty.ai@nycu.edu.tw,\textsuperscript{\rm 2}jwhsieh@nycu.edu.tw,\textsuperscript{\rm 3}yctseng@cs.nycu.edu.tw,\textsuperscript{\rm 4}mingching@gmail.com}
\maketitle

\begin{abstract}
We develop a Synthetic Fusion Pyramid Network (SPF-Net) with a scale-aware loss function design for accurate crowd counting. Existing crowd-counting methods assume that the training annotation points were accurate and thus ignore the fact that noisy annotations can lead to large model-learning bias and counting error, especially for counting highly dense crowds that appear far away. To the best of our knowledge, this work is the first to properly handle such noise at multiple scales in end-to-end loss design and thus push the crowd counting state-of-the-art. We model the noise of crowd annotation points as a Gaussian and derive the crowd probability density map from the input image. We then approximate the joint distribution of crowd density maps with the full covariance of multiple scales and derive a low-rank approximation for tractability and efficient implementation. The derived scale-aware loss function is used to train the SPF-Net. We show that it outperforms various loss functions on four public datasets: UCF-QNRF, UCF CC 50, NWPU and ShanghaiTech A-B datasets. The proposed SPF-Net can accurately predict the locations of people in the crowd, despite training on noisy training annotations.


\end{abstract}

\section{Introduction}

\begin{figure}[t]
\centerline{
  \includegraphics[width=0.49\textwidth]{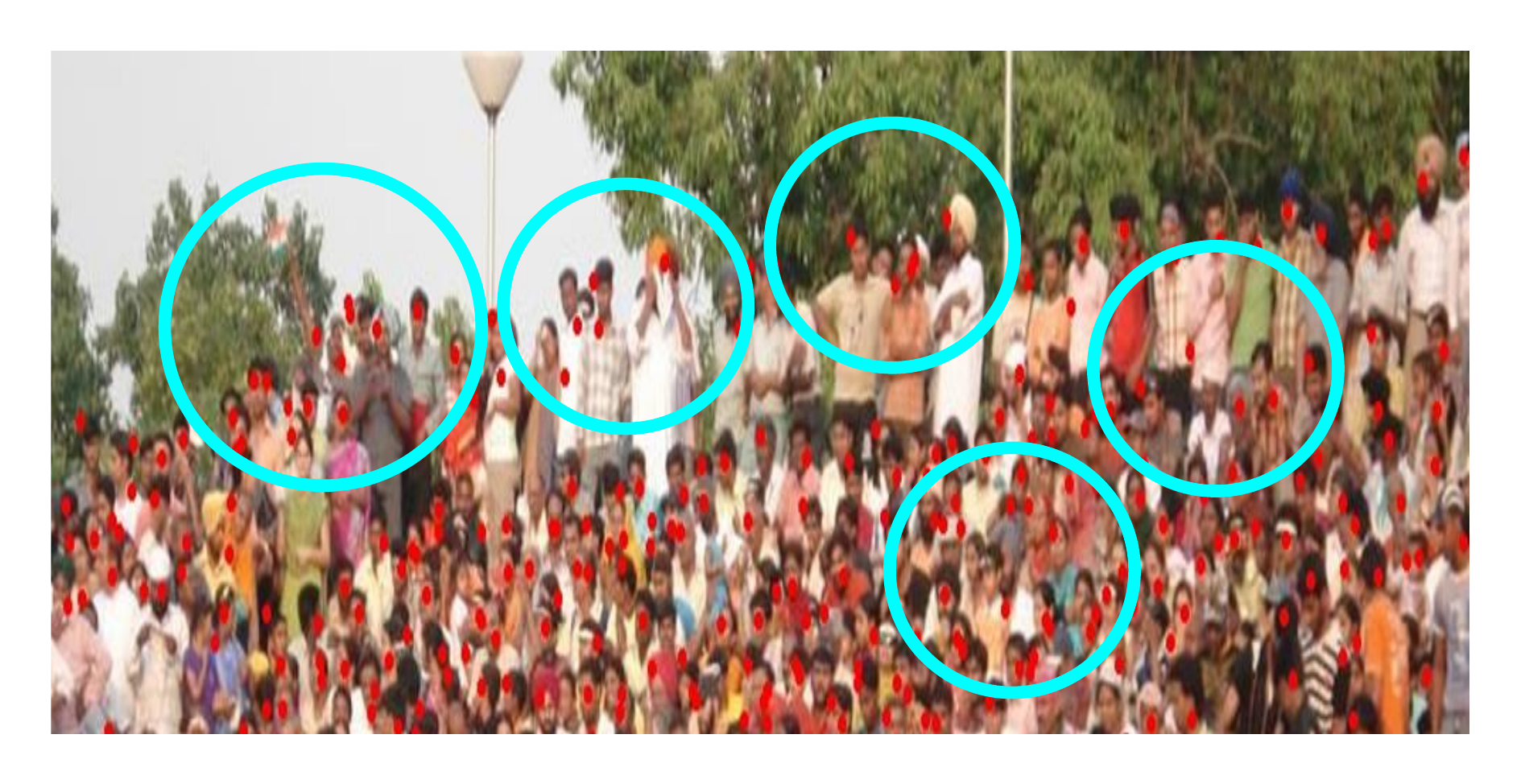}
  \vspace{-0.5cm}
}
\caption{An example of crowd counting ground truth showing various inaccurate annotations in the ShanghaiTech A dataset~\cite{zhang2016single}.
}
\label{fig:annotation noise}
\vspace{-0.3cm}
\end{figure}

Crowd counting is an emerging technology in computer vision that is useful for public safety and crowd behavior analysis. Many CNN-based crowd-counting methods have been developed over the years~\cite{li2018csrnet,xu2019autoscale,bai2020adaptive, ma2019bayesian,xiong2019open,varior2019multi,jiang2020attention,thanasutives2021encoder, zhu2019dual}, based on the prediction of crowd density map from a given image, where the sum of density maps in the image is the total number of people predicted. Many methods cast this map prediction problem as a standard regression task using the standard L2 norm~\cite{li2018csrnet,wan2019adaptive,cao2018scale} or Bayesian Loss (BL)~\cite{ma2019bayesian} as the loss function.  The loss function measures the difference between the estimated density map and the ground truth annotation of the human heads of the crowd. A major drawback of existing crowd counting methods that use L2 or BL loss is that the model training process is based on an important assumption of {\em perfect annotation,} {\em that is}, precise annotation of head positions without error. However, inevitably, annotation noise or error is present during the groundtruth-labeling process. It is difficult for even a human annotation to precisely localize the center of the head of each person in an image, especially for people that appear small or far away, as shown in Figure~\ref{fig:annotation noise}. An important motivation of this work is that we hypothesize such annotation error can be modeled using a carefully designed loss function and then corrected to generate an improved crowd-counting heat map. Furthermore, in existing methods, these two loss functions assume that each pixel is {\em independent and identically distributed} (i.i.d.). When two heads are highly overlapped or occluded, nearby pixels in the density map are not longer i.i.d.

Noise in human annotations, $i.e.$, displacement of annotation locations, and the correlation between pixels should be considered in the modeling process to achieve highly accurate crowd counting. Wang and Chan \cite{wan2020modeling} used a multivariate Gaussian distribution to deal with the problems of annotation noise and correlation.  In order to speed up the efficiency of crowd counting, a low-rank method is adopted to approximate the covariance matrix of this Gaussian distribution for efficient and tractable implementation.  However, the following problem occurs where \cite{wan2020modeling} did not consider the scaling problem.  The sizes of human heads will not be the same at any positions in the image, so the covariance matrix used for annotation correction cannot be fixed for all objects with different sizes.  It should be sensitive to scale and adapt to changes in head size, as shown in Figure~\ref{fig:annotation noise}.  

To address the above issues, we propose a novel {\bf scale-aware loss function} that simultaneously considers annotation noise, head-to-head correlation, and adjustment for variances at different scales. First, we derive the form of the marginal probability distribution of the population density values at each head location $x$ on the density map. The density values of the position $x$ at different scales are approximated by a Gaussian distribution. In order to model the correlation between pixels at different scales, we derive the multivariate Gaussian distribution with the full covariance matrix of different scales. Since the time complexity to calculate the multivariate Gaussian is equal to the square of image size, a low-rank approximation method is adopted to speed up the efficiency of crowd counting.  Finally, a novel scale-aware loss function is defined to model human annotation errors and the correlations between heads to train the density map estimator more accurately.

Another novelty of this paper is that we develop a new architecture to effectively improve the counting accuracy of small objects.  The Feature Pyramid (FP) can capture an object's visual features at different scales and thus significantly improve the accuracy of object counting. Due to this increase in accuracy, FP structure has become the standard component for most SoTA object counting frameworks~\cite{li2018csrnet,xu2019autoscale,bai2020adaptive, ma2019bayesian,xiong2019open,varior2019multi,jiang2020attention,thanasutives2021encoder, zhu2019dual}. However, the feature maps in this FP structure are scaled to $1/2$, $1/4$, or $1/8$ of the input size, where the scale gap and truncation will cause the features of small objects to disappear dramatically.  We address this problem by adding a synthetic fusion layer to this FP structure. Specifically, we enable the synthetic fusion layer to scale the density map to $1/2$, $1/3$, $1/4$, or $1/6$. This fusion layer is for reducing the effects of scaling truncation. We believe that the inclusion of the middle-scale feature map makes the transition of different scales smoother for better crowd counting.

\begin{figure*}[t]
\centerline{
\includegraphics[width=0.96\linewidth]{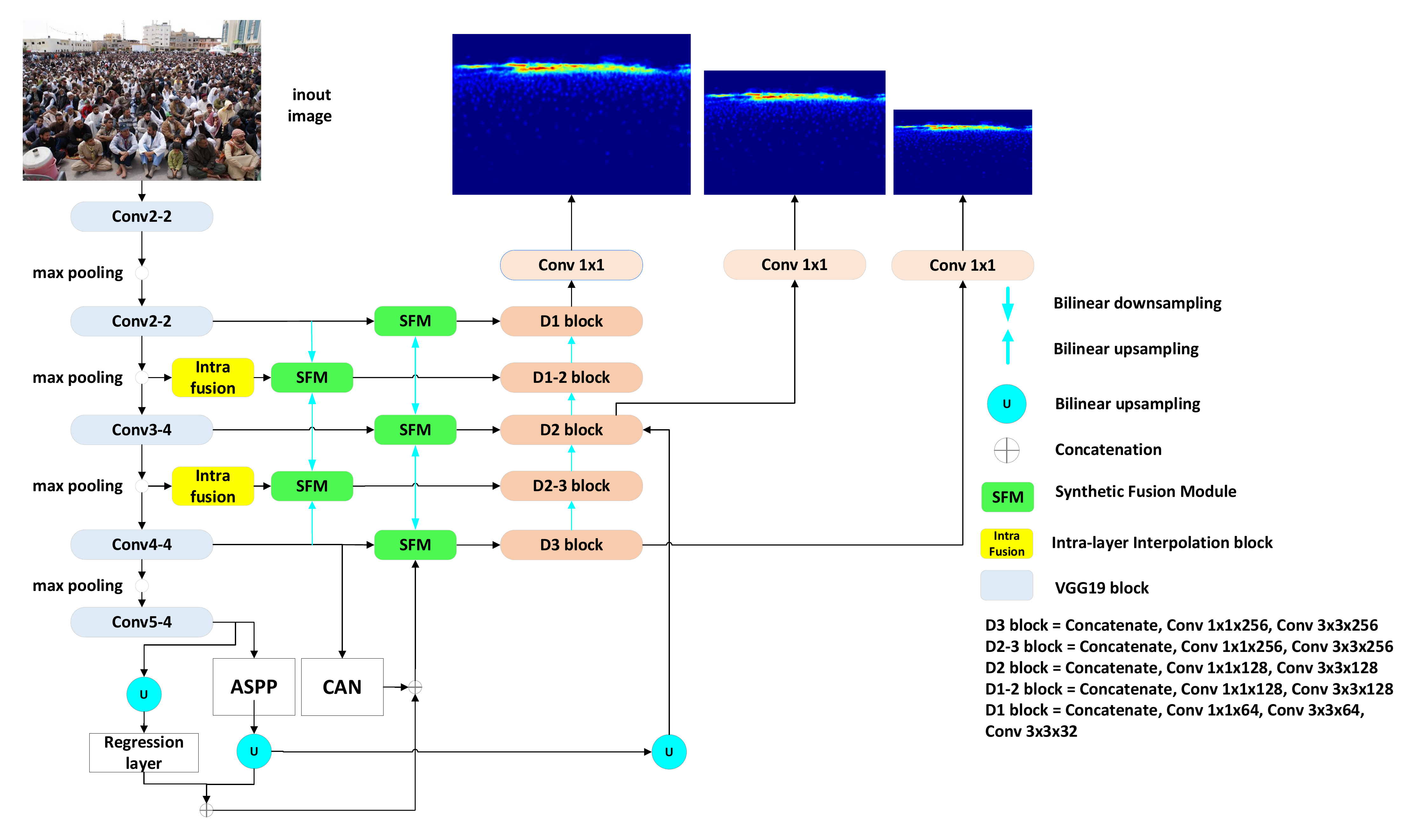}
\vspace{-0.3cm}
}
\caption{The proposed Synthetic Fusion Pyramid Network (SFP-Net) architecture for scale-aware crown counting. With a new scale-aware loss function, it outperforms all SoTA methods on four main public crowd-counting datasets.}
\label{fig:Scale-IFCnet}
\vspace{-0.6cm}
\end{figure*}

Our new architecture, {\bf Synthetic Fusion Pyramid Network (SFP-Net)} is integrated into VGG-19 and trained by the new scale-aware loss to achieve SoTA performance on the UCF-QNRF~\cite{idrees2018composition}, NWPU~\cite{Wang2020NWPU}, UCF CC 50\cite{idrees2013multi}, and ShanghaiTech A-B~\cite{zhang2016single} datasets.  Our MAE=78.36 on the UCF-QNRF dataset is currently the only method below 80.  Main contributions of this paper are summarized as follows:

\begin{itemize}[leftmargin=16pt] \itemsep -.5em
\item We develop a new loss function to model annotation noise and the head-to-head correction with a scale-aware multivariate Gaussian distribution so that better density maps can be generated for effective crowd counting. 

\item A new Synthetic Fusion Pyramid Network (SFP-Net) is proposed to generate a fine-grained feature map to count small objects more effectively and accurately.

\item The proposed SFP-Net with the new scale-aware loss function outperforms all SoTA methods on four main public crowd-counting datasets.  
\end{itemize}

\section{Related Works}

In the literature, there are several frameworks such as \cite{gall2011hough,viola2005detecting} based on hand-crafted features to count objects with low densities. These methods cannot handle cases with diverse crowd distributions. Inspired by the success of heatmap-based object detection, many CNN architectures~\cite{li2018csrnet,xu2019autoscale,bai2020adaptive,xiong2019open,varior2019multi,jiang2020attention,thanasutives2021encoder, zhu2019dual} are proposed to use various regression models to improve density maps, where the sum of the density map yields the crowd count. Details of these architectures are surveyed below.

{\bf Handle scale variations in crowd counting.}
One critical challenge of crowd counting based on the sum of density map is the scale variation due to various distances between the viewing cameras and the targets.  In the multicolumn CNN of \cite{zhang2016single}, each column uses a different combination of convolution kernels to extract multiple-scale features. However, the results of \cite{li2018csrnet} show that similar features are often learned in each column of this network; and thus the model cannot be efficiently trained as the layers go deeper. In \cite{li2018csrnet}, multi-scale features are obtained using VGG16 and convolutions are adopted with different dilation rates. Instead of using different conv kernel sizes in each layer, a multi-branch strategy is used in \cite{varior2019multi} to choose convolution filters with a fixed size to extract multiple-scale features across layers. To avoid repeatedly computing convolutional features from subregions, in \cite{xiong2019open}, the multiresolution feature maps are generated by dividing a dense region into subregions, whose counts are calculated within a previously observed closed set. In \cite{liu2019context}, scale variation is handled by encoding multi-scale contextual information into the regression model. In \cite{jiang2020attention}, a density attention network generates various attention masks to focus the task of crowd counting on a particular scale. A densely connected architecture is used in \cite{miao2020shallow} to maintain multi-scale information well, while an attention mechanism is used to remove the background noise. In \cite{thanasutives2021encoder}, the CAN~\cite{liu2019context} and ASPP~\cite{chen2017deeplab} are integrated into the regression model to obtain multi-scale features for better heat map generation.  To improve generalizability, \cite{zhang2015cross} proposed a CNN architecture based on a switching strategy to perform an alternative optimization between density estimation and count estimation. 


{\bf Deviation and errors from point-wise annotations.} 
In most crowd-counting datasets, a dotted annotation is often adopted to represent each object in images and profoundly affects subjective deviation and
performance evaluation compared to the bounding-box annotation since the dotted annotation does not include
any size information.  To address this issue, in \cite{zhang2016single}, the average distance from each head to its three neighbors is calculated, and then the head size is estimated as the Gaussian standard deviation.  In \cite{cheng2022rethinking}, various locally connected Gaussian kernels are used to replace the original convolution filter to reduce noise during the density map generation process.

{\bf Loss Function:} In the literature, the pixel-wise Mean Square Error (MSE) loss has dominated the training of density estimation-based crowd counting approaches.  In~\cite{jiang2019crowd}, the annotation deviation is reduced by using a combinatorial loss that includes a spatial abstraction term and a spatial correlation term. The Bayesian loss is used in \cite{ma2019bayesian} to construct a density contribution probability model to reduce the influence of deviation. However, this approach cannot handle false positives well. Furthermore, a DM-count loss is defined in \cite{wang2020distribution} to measure the similarity between the normalized predicted density map and the ground truth.  In \cite{wan2020modeling}, a loss function with a multivariate Gaussian distribution is defined to deal with the problems of annotation noise and correlation.  However, this loss is not scale-aware. In general, a pixel shift in annotation might not affect the accuracy of large object counting; however, it will result in significant errors in small object counting. 

\section{Proposed Method and Architecture}
To better understand our proposed scale-aware loss function, we first review the traditional density map generation and then describe the scaling effect of the label noise on the heat map generation.  After deriving our scale-aware loss, the final SPF-Net is proposed for crowd counting.

\subsection{Background: Density Map Generation}
\label{sec:DM}

Traditional mainstream methods turn the counting task into a density regression problem \cite{lempitsky2010learning,Sindagi_2017_ICCV, wan2020modeling}. For $N$ annotations of an image $\mathcal{I}$, the annotation points of all heads are $\{\tilde{\mathrm{\textbf{H}}}_{i}\}^{N}_{i=1}$  where $\tilde{\mathrm{\textbf{H}}}_{i}$ corresponds to the position of the $i$th head annotated in $\mathcal{I}$ with noise labeling. For a given spatial location $x$ in $\mathcal{I}$, obtain the corresponding density value $y$ at the location of the pixel $x$ by placing a Gaussian kernel in each annotation,
\begin{equation}
    y(x) = \sum_{i=1}^{N}\mathcal{N}(x|\tilde{\mathrm{\textbf{H}}}_{i},\beta\mathrm{\textbf{I}})=\sum_{i=1}^{N}\frac{1}{\sqrt{2\pi}\beta}exp(-\frac{|| x-\tilde{\mathrm{\textbf{H}}}_{i} ||_{\beta\mathrm{\textbf{I}}}^{2}}{2}),
\label{eq:gau ker}
\end{equation}
where $\beta$ is the variance of the Gaussian kernel and $\sum_{i=1}^{N}\mathcal{N}(x|\mu,	\Sigma)$ is the Probability Density Function (PDF) for a multivariate Gaussian with mean $\mu$ and covariance matrix $\Sigma$,  with $\left \| x \right \|^{2}_{\Sigma }=X^{T}\Sigma ^{-1}X$ as the squared Mahalanobis distance and $X$ is the feature vector of $x$ extracted from a backbone.
Through Eq.(\ref{eq:gau ker}) for all head positions in the image, the density map $y$  is estimated from $\mathcal{I}$ by learning a regressor $f(\mathcal{I})$ based on the L2-norm loss $\mathcal{L}(y,f(\mathcal{I}))$ = $\left\| y-f(\mathcal{I}) \right\|^{2}$ or a Bayesian loss~\cite{wan2020modeling}.  Then, the sum of all pixels in $y$ is the crowd count.  However, as shown in Figure~\ref{fig:annotation noise}, the ground truth includes various annotation errors.  Thus, $y$ is often wrongly estimated by not only observation noise but also the above annotation errors.  In what follows, a new scale-aware loss function is proposed to deal with this problem of annotation noise.

\subsection{Scale-Aware Annotation Noise}
 Let $\mathrm{\textbf{H}}_{i}$ be the true location of the $i$-th head, and $ \tilde{\mathrm{\textbf{H}}}_{i}=\mathrm{\textbf{H}}_{i}+\varepsilon_{i}$,  where $\varepsilon_{i}$ is the annotation noise. The annotation noise $\varepsilon_{i}\overset{i.i.d}{\sim }\mathcal{N}(0,\alpha\mathit{\textbf{I}})$, where $\alpha$ is the variance parameter. The density value at the location $x$ is modeled as follows:
\begin{equation}
\begin{aligned}
    \mathbb{D}(x) &=  \sum_{i=1}^{N}\mathcal{N}(x|\mathrm{\tilde{\mathrm{\textbf{H}}}_{i}},\beta\mathit{\textbf{I}})=\sum_{i=1}^{N}\mathcal{N}(x|{\mathrm{\textbf{H}}}_{i}+\varepsilon_{i},\beta\mathit{\textbf{I}})\\
    &=\sum_{i=1}^{N}\mathcal{N}(q_{i}|\varepsilon_{i},\beta\textbf{I})\cong \sum_{i=1}^{N}\phi_{i}.
    \end{aligned}
    \label{eq:density}
\end{equation}
where $\mathbb{D}(x)$ is the density of the location $x$, $\phi_{i}$ is the individual term for the $i$th annotation and $q_{i}=x-{\mathrm{\textbf{H}}}_{i}$, the difference between the location of the $i$-th annotation and the location $x$. In fact, the ranges of annotation error are similar for large objects and small objects and lead to a scale-fixed model used in ~\cite{wan2020modeling} for modeling annotation noise.  However, a one-pixel shift error in annotation might not cause problems when counting large objects, but will result in a significant accuracy reduction when counting small objects.  Thus, modeling the annotation noise should be scale-aware.      
\begin{figure}[t]
\centerline{
  \vspace{-0.3cm}
  \includegraphics[width=0.5\textwidth]{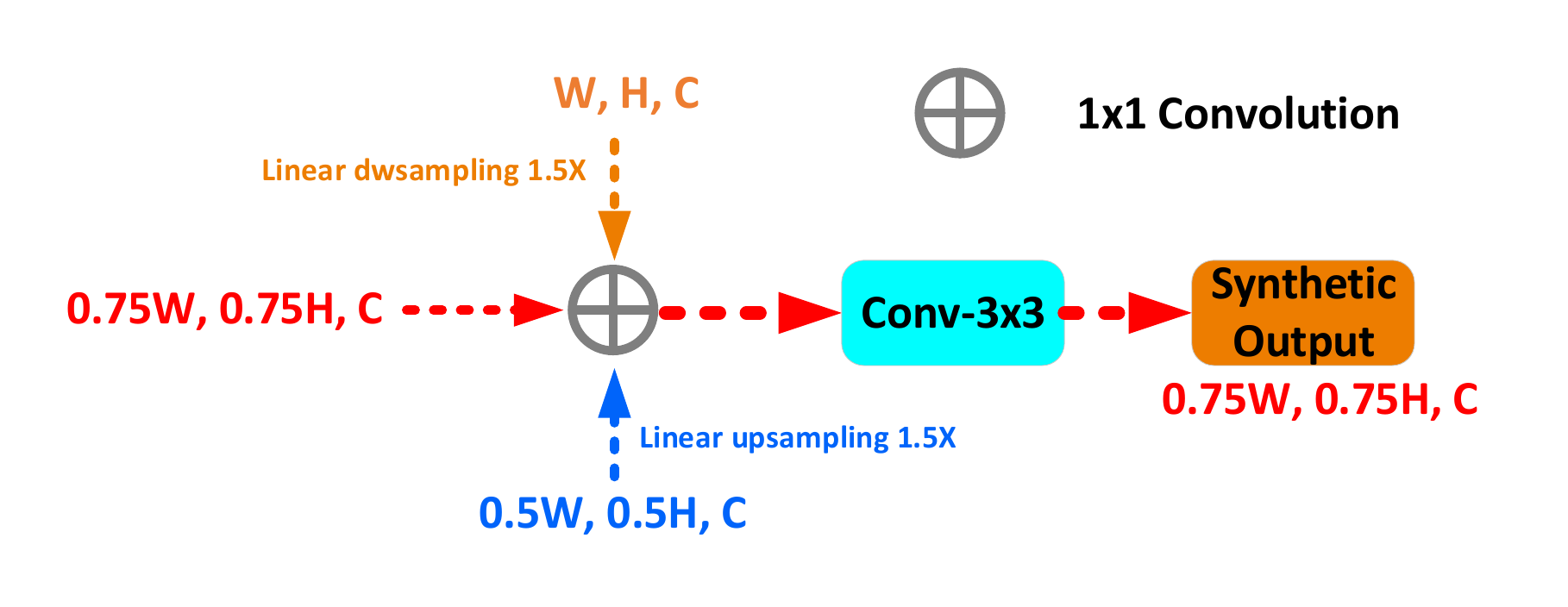}
  \vspace{-0.3cm}
}
\caption{Synthetic Fusion Module.
\vspace{-0.2cm}
}
\label{fig:SFM}
\end{figure}
Assume that there are $S$ scales used to model the annotation noise.  Then, Eq.(\ref{eq:density}) can be rewritten as

\begin{equation}
\begin{aligned}
    \mathbb{D}(x) &= \sum_{s=1}^{S}  \mathbb{D}_{s}(x)=\sum_{s=1}^{S} \sum_{i=1}^{N}\mathcal{N}(q_{i}|\varepsilon_{i},\beta_s\textbf{I})\cong \sum_{s=1}^{S} \sum_{i=1}^{N}\phi_{i}^s,
    \end{aligned}
    \label{eq:all-density}
\end{equation}
where  $ \phi_i^s=\mathcal{N}(q_{i}|\varepsilon_{i},\beta_{s}\textbf{I}) $, $i.e$., the Gaussian kernel placed in the $i$th annotation at the scale $s$ and parameterized with the annotation error $\varepsilon_{i}$ and the variance $\beta_s$.  In addition, we set $\beta_{s+1}=\beta_s/2$ since a pooling operation will cause the feature map to be scaled by 2.  Let $\mathcal{I}_s$ be the scaled-down version of $\mathcal{I}$ at the scale $s$. For all pixels $x_j$ in $\mathcal{I}_s$, a multivariate random variable for the density map $\mathbb{D}(x)$ can be constructed as $\textbf{D}_{s}=[\mathbb{D}_{s}(x_1), \cdots, \mathbb{D}_{s}(x_j), \cdots , \mathbb{D}_{s}(x_{J_s})]$, where $J_s$ is the number of pixels in $\mathcal{I}_s$. 

\subsubsection{Scale-aware Probability Distribution}
To calculate the sum 
$\mathbb{D}_{s}= \sum_{i=1}^{N}\phi_{i}^s$ in closed form, we approximate it by a Gaussian, $i.e$., $\hat{p}(\mathbb{D}_{s})\sim\mathcal{N}(\mathbb{D}_{s}|\mu_{s}, \sigma^{2}_{s})$ with the scale-aware mean $\mu_{s}$ and variance $\sigma^{2}_{s}$.  The mean $\mu_{s}$ is calculated as follows:  
\begin{equation}
\begin{aligned}
    \mu_{s}&= \mathbb{E}[\mathbb{D}_{s}]=\mathbb{E}[ \sum_{i=1}^{N}\mathcal{N}(q_{i}|\varepsilon_{i},\beta_{s}\textbf{I})] \\ &
    =\sum_{i=1}^{N}\mathcal{N}(q_{i}|0,(\alpha+\beta_{s})\textbf{I})\cong \sum_{i=1}^{N}\mu_i^s,
    \label{eq:mu}
\end{aligned}
\end{equation}
where the annotation error $\varepsilon_{i}\sim\mathcal{N}(0|0, \alpha \textbf{I})$. The variance $\sigma^{2}_{s}$ is given by
\begin{equation}
\begin{aligned}
    \sigma^{2}_{s}&=\mathrm{var}(\mathbb{D}_{s})=\mathbb{E}[\mathbb{D}_{s}^{2}]-\mathbb{E}[\mathbb{D}_{s}]^{2}\\ &\cong\sum_{i=1}^{N}\frac{1}{4\pi\beta_{s}}\mathcal{N}(q_{i}|0,(\beta_{s}/2+\alpha)\textbf{I})-\sum_{i=1}^{N}(\mu_i^s)^{2}.
    \label{eq:var}
\end{aligned}
\end{equation}



\begin{figure}[t]
\centerline{

  {\footnotesize (a)} 
  \includegraphics[width=0.18\textwidth]{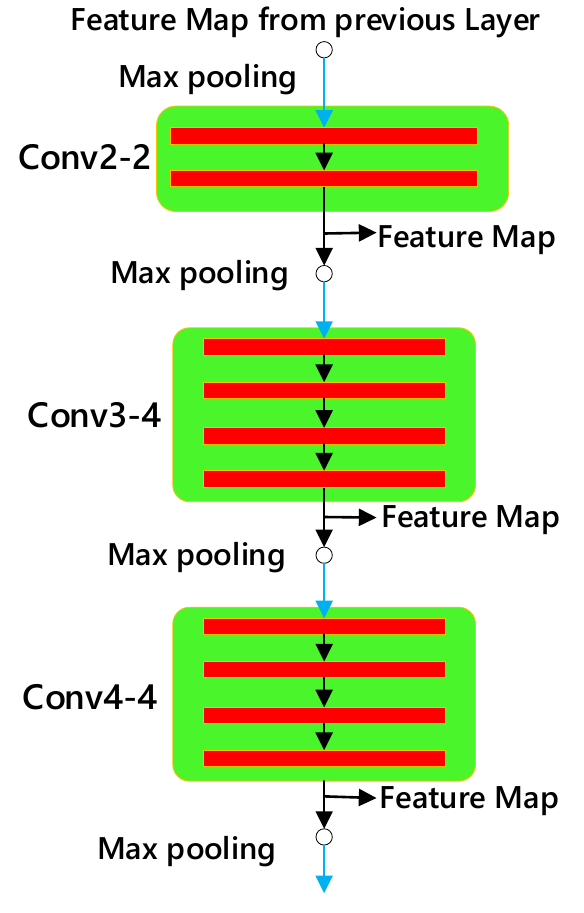}
    \vspace{-0.1cm}
  {\footnotesize (b)}
  \includegraphics[width=0.25\textwidth]{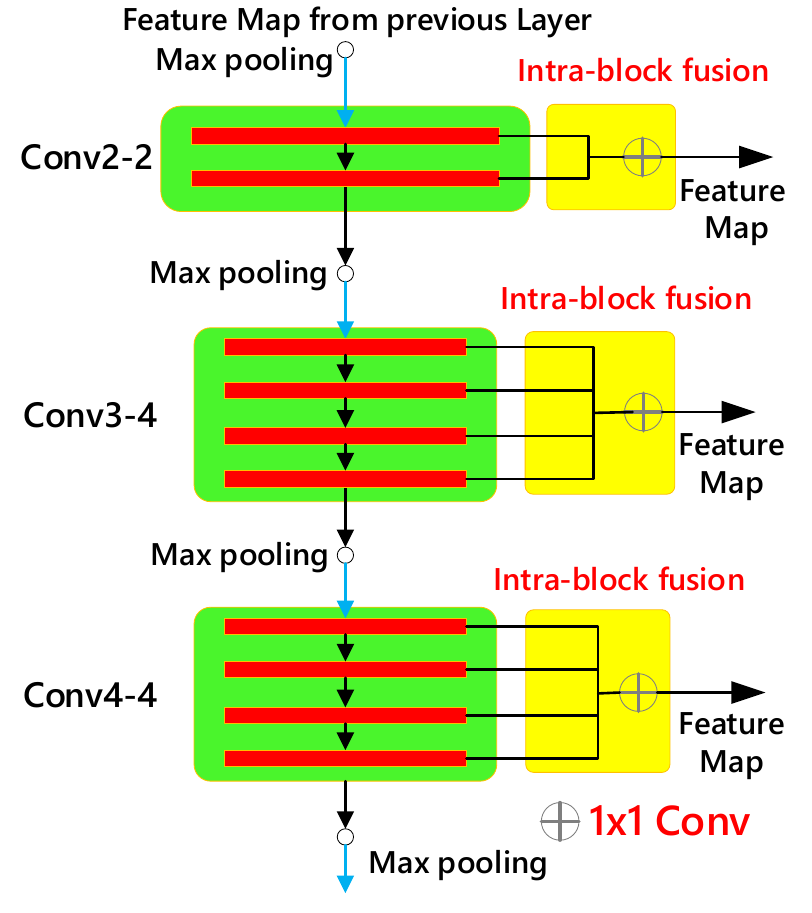}
  \vspace{-0.1cm}
}
\caption{
Without and with Intra-block Fusion.
(a) Without Intra-block Fusion.
(b) Intra-block Fusion.
\vspace{-0.5cm}
}
\label{fig:intra_block_fusion}
\end{figure}

\subsubsection{Gaussian Approximation to Scale-Aware Joint Likelihood $\textbf{D}_{s}$}
Next, we consider the correlation between locations and model it by a multivariate Gaussian approximation of the joint likelihood $\textbf{D}_{s}$ at different scales $s$. Let $q_{i}{(x_j)}=x_j-\tilde{\mathrm{\textbf{H}}}_{i}$ be the difference between the spatial location of the $i$-th annotation and the location of the pixel $x_j$. Based on Eq.(\ref{eq:all-density}), the density value $\mathbb{D}_{s}{(x_j)}$ is calculated as 
\begin{equation}
    \mathbb{D}_{s}{(x_j)} = \sum_{i=1}^{N}\mathcal{N}(q_{i}{(x_j)}|\varepsilon_{i},\beta_s\textbf{I})\cong \sum_{i=1}^{N}\phi_i^s{(x_j)}.
    \label{eq:Phi}
\end{equation}
Note that the annotation noise $\varepsilon_{i}$ is the same random variable across all $\phi_i^s{(x_j)}$.  Define the Gaussian approximation to $\textbf{D}_{s}$ as $\hat{p}(\textbf{D}_{s})$ = $\mathcal{N}(\textbf{D}_{s}|\mu_{s},\Sigma_{s})$, where $\mu_{s}$ and $\Sigma_{s}$ are defined in Eq.(\ref{eq:mu}) and Eq.(\ref{eq:var}), respectively. The $j$th entry in $\mu_{s}$ is $\mathbb{E}[\mathbb{D}_{s}(x_j)]$ = $\sum_{i=1}^{N}\mu_i^s{(x_j)}$ (see Eq.(\ref{eq:mu})) and the diagonal of the scale-aware covariance matrix is calculated as $\boldsymbol{\Sigma}_{x_j,x_j}^s$ = $\mathrm{Var}(\mathbb{D}_{s}{(x_j)})$. The covariance term is derived as
\begin{equation}
\begin{aligned}
    \boldsymbol{\Sigma}_{x_j,x_k}^s &= \mathrm{Cov}(\mathbb{D}_{s}{(x_j)},\mathbb{D}_{s}{(x_k)})\\
    &= \sum_{i=1}^{N}\omega_i^s{(x_j,x_k)}-\sum_{i=1}^{N}\mu_i^s{(x_j)}\mu_i^s{(x_k)}.
    \end{aligned}
    \label{eq:cov}
\end{equation}
where $\omega_i^s{(x_j,x_k)}$ = $\mathbb{E}[\phi_i^s{(x_j)}\phi_i^s{(x_k})]$.

\subsection{Scale-Award Covariance Matrix with Low-rank Approximation}
Since the dimension of $\boldsymbol{\Sigma}_{x_j,x_k}^s$ is huge, $i.e$, $J_s \times J_s$, this section derives its low-rank approximation with its nonzero rows and columns for efficiency improvement.  Let $v_j^s$ = $\boldsymbol{\Sigma}_{x_j,x_j}^s$.  To obtain this low-rank approximation, each pixel $x_j$ is first ordered by $v_j^s$. Then, the top-$M$ pixels whose percentages of variance are larger than 0.8, $i.e$. 
\begin{equation}
  \frac{\sum_{j=1}^{M}v_j^s}{\sum_{i=1}^{J} v_j^s} > 0.8,
\end{equation}
are selected from $\mathcal{I}_s$ for this low-rank approximation. Let the set of indices of the top $M$ pixels be denoted by $L$, $i.e.$, $L=\{l_1,l_2,\dots,l_m,\dots,l_{M}\}$. Then, only the elements of $L$ are selected to approximate $\boldsymbol{\Sigma}^{s}$.  Let $\hat{\boldsymbol{\Sigma} }^{s}$ denote the approximation to $\boldsymbol{\Sigma}^{s}$ and be calculated as follows.

\begin{equation}
  \hat{\boldsymbol{\Sigma} }^{s} \cong \boldsymbol{\Sigma}^{s}  = \textbf{V}^{s} +\textbf{M}\textbf{C}_{L}^{s}\textbf{M}^{T},
    \label{eq:Sigma}
\end{equation}
where $\textbf{V}^s$ is the $J_s\times J_s$ diagonal matrix formed by the diagonal of $\boldsymbol{\Sigma}^{s}$ and $\textbf{C}_{L}^{s}$ are the $J_s\times J_s$ covariance matrix and $\textbf{M}$ is a $J_s\times M$ permutation matrix whose $m$-th column $[\textbf{M}]_{m}=e_{l_m}$, where $e_i$ is the $i$-th Canonical unit vector. In addition,  $\textbf{C}_{L}^{s}$ is obtained by
\begin{equation}
\textbf{C}_{L}^{s}(i,j)=\left\{ {\begin{array}{*{2}{c}}
{0\ \quad\quad\quad\quad\quad\quad\quad\quad,~{\rm{  if }} ~i=j,}\\
{\mathrm{Cov}(\mathbb{D}_{s}{(x_{l_i})},\mathbb{D}_{s}{(x_{l_j})}),~ {\rm{  if }}~ i \ne j.}
\end{array}} \right.
\end{equation}
Using the matrix inversion lemma, the approximate inverse covariance matrix $(\hat{\boldsymbol{\Sigma}}^{s})^{-1}$ can be obtained as follows:
\begin{equation}
  (\hat{\boldsymbol{\Sigma}}^{s})^{-1}=(\textbf{V}^{s})^{-1} - \textbf{M}\textbf{B}_{L}^{s}\textbf{M}^{T},
    \label{eq:InverseSigma}
\end{equation}
 where $\textbf{B}_{L}^{s}=\textbf{V}_{L}^{s}(\textbf{C}_{L}^{s})^{-1} \textbf{V}_{L}^{s} + \textbf{V}_{L}^{s}$ and $\textbf{V}_{L}^{s}=\textbf{M}^{T}\textbf{V}^{s}\textbf{M}$. Finally, the approximate negative log-likelihood function is
\begin{equation}
\begin{aligned}
    -\mathrm{log} \hat{p}(\textbf{D}_{s}) &= -\mathrm{log} \mathcal{N}(\textbf{D}_{s}|\mu_{s},\hat{\boldsymbol{\Sigma}}^{s})\propto ||\textbf{D}_{s}-\mu_{s}||^{2}_{\hat{\boldsymbol{\Sigma}}^{s}},
    \label{eq:apro_likelhood}
    \end{aligned}
\end{equation}
where 
\begin{equation}
\begin{aligned}
   ||\textbf{D}_{s}-\mu_{s}||^{2}_{\hat{\boldsymbol{\Sigma}}^{s}} &= \bar{\textbf{D}}_{s}^{T}(\textbf{V}^{s})^{-1}\bar{\textbf{D}}_{s}-\bar{\textbf{D}}_{s}^{T}\textbf{MB}_{L}^{s} \textbf{M}^{T}\bar{\textbf{D}}_{s},
    \label{eq:apro Phi}
    \end{aligned}
\end{equation}
and $\bar{\textbf{D}}_{s}=\textbf{D}_s-\mu_{s}$. The correlation term in Eq.(\ref{eq:apro Phi}) is based on the set $L$. Thus, the storage / computation complexity of one training example using the low-rank approximation is $O(M^{2}+J_s)$ compared to $O(J_s^{2})$ for the full covariance.

\subsection{Regularization and the Final Loss Term}

 To ensure that the predicted density map near each annotation satisfies the density sum to 1, for the $i$-th annotation point, we define the regularizer $\mathcal{R}_{i}^{s}$ as follows: 
\begin{equation}
    \mathcal{R}_{i}^{s} = |\sum_{j}D_{s}(x_j)\frac{p(\phi_{i}^s(x_j))}{\sum_{i=1}^{N}p(\phi_{i}^s{(x_j)})}-1|.
\end{equation}
Then, the final loss function is defined as:
\begin{equation}
    \mathcal{L} =  \sum_{s=1}^{S}w_{s}\bar{D}_{s}^{T}(\hat{\Sigma}^{s})^{-1}\bar{D}_{s}+\sum_{s=1}^{S}\sum_{i=1}^{N} \mathcal{R}_{i}^{s}, 
\end{equation}
where $w_s$ is the weight of the scale $s$ and $\sum_{s=1}^{S}w_{s}=1$.

\subsection{The Synthetic Fusion Pyramid Network}
\label{sec:SFPnet}

After deriving the scale-aware loss function, a new architecture is proposed to achieve SoTA accuracy in crowd counting.  Figure~\ref{fig:Scale-IFCnet} shows the architecture of the proposed {\bf Synthetic Fusion Pyramid Network (SFP-Net)} for crowd computing.  This SFP-Net adopts VGG19 as the backbone to extract various features.  However, in most of CNN backbones, the used pooling operation (or convolution with stride 2) usually down-samples the image dimension to half and makes the density map scaled to 1/2, 1/4, 1/8, and so on in both the $x$ and $y$ directions. We believe that the scale gap is too large and causes the features fusion of layers to be uneven.  One novelty of this paper is a Synthetic Fusion Module (SFM) proposed to generate various synthetic layers between the original layers so that better density maps can be constructed for crowd counting. In addition, an Intra-block Fusion Module (IFM) is proposed to allow all feature layers within the same convolution block to be fused so that more fine-grained information can be sent to the decoder for more effective crowd counting.  At the last layer, the ASPP \cite{chen2017deeplab} and CAN \cite{liu2019context} modules are adopted to use atrous convolutions with different rates to extract multiscale features for counting objects more accurately.  This SFP-Net with the new scale-aware loss function achieves state-of-the-art performance on the UCF-QNRF~\cite{idrees2018composition}, NWPU~\cite{Wang2020NWPU}, UCF CC 50\cite{idrees2013multi}, and ShanghaiTech A-B~\cite{zhang2016single} datasets.  Details of SFM and IFM are described as follows.

{\bf Synthetic Fusion Module:} The SFM creates various synthetic layers between the original layers to make the prediction maps scaled to 1/2, 1/3, 1/4, 1/6 and 1/8. Thus, a smoother scale space is provided for fitting the ground truth whose scale changes continuously.  As shown in Figure~\ref{fig:SFM}, the synthetic layer is generated by an SFM (denoted green color) that can have two or three inputs, depending on its position in the SFP-Net.  SFM performs by first linearly scaling inputs, then merging them via 1$\times$1 convolution, and finally fusion with a conv-3$\times$3. It can synthesize the original/synthetic layers or simply fuse features.

{\bf Intra-block Fusion Module:} Figure~\ref{fig:intra_block_fusion}(a) shows a convolution block in a backbone, there are various convolution layers used to generate a feature map. Figure~\ref{fig:intra_block_fusion}(b) shows that, a new Intra-block Fusion Module (IFM) is adopted to fuse all convoluted results within the same convolution block through $1\times1$ to generate more fine-granted details to the feature map, which is then sent to the decoder to construct a better heat map for crowd counting.

\section{Experimental Results}  

We evaluated our crowd-counting method and compared it with eleven SoTA methods in four public datasets, UCF-QNRF~\cite{idrees2018composition}, UCF CC 50~\cite{idrees2013multi}, NWPU-Crowd~\cite{Wang2020NWPU}, and ShanghaiTech Parts A and B~\cite{zhang2016single}.

\subsection{Data Sets}

The {\bf UCF-QNRF} dataset~\cite{idrees2018composition} contains $1,535$ crowded images with average resolution $2,013 \times 2,902$.  It is the most challenging dataset with very large crowds, wider varieties of scenes, viewpoints, densities, lighting images, and $1,251,642$ annotated people. The training and test sets include $1,201$ and $334$ images, respectively.

The {\bf NWPU-Crowd} dataset~\cite{Wang2020NWPU} contains $5,109$ images and $2,133,375$ annotated instances with point and box labels.  The dataset is divided into $3,109$ training images, $500$ validation images, and $1,500$ test images.  It is the largest dataset whose testing labels are not released.

The {\bf ShanghaiTech} dataset \cite{zhang2016single} is divided into two parts A and B. Part A contains $482$ images with an average resolution $589 \times 868$. The numbers of people vary from $33$ to $3,139$ with annotated people $241,677$.  Part B contains $716$ images with crowd numbers ranging from $9$ to $578$. 

The {\bf UCF CC 50} dataset \cite{idrees2013multi} contains 50 gray images with different resolutions. The average count for each image is 1,280, and the minimum and maximum counts are 94 and 4,532, respectively. Since this is a small-scale dataset and no data split is defined for training and testing, we perform five-fold cross-validations to get the average result.

\subsection{Model Training Parameters.}

Our method is trained from parameters pre-trained on ImageNet~\cite{imagenet} with the Adam optimizer.  Since the image dimensions in the used datasets are different, patches with a fixed size are cropped at random locations, then randomly flipped (horizontally) with probability 0.5 for data augmentation.  The learning rates used in the training process are $1e^{-5}$, $1e^{-5}$, $1e^{-5}$, and $1e^{-4}$ for the UCF-QNRF, UCF CC 50, NWPU, and ShanghaiTech datasets, respectively.   To stabilize the training loss change, we use batch sizes 10, 10, 15, and 10, respectively.  All parameters used in the training procedure are listed in Table~\ref{tab:Training_p}.

\begin{table}[t]
\caption{Detailed parameters used for training.}
\begin{tabular}{c|c|c|c}
\hline
Dataset      & learning rate & batch size & crop size       \\ \hline
UCF-QRNF     & 1e-5     & 12       & 512×512 \\
UCF CC 50    & 1e-5     & 10       & 512×512 \\
NWPU         & 1e-5     & 8       & 512×512 \\
ShanghaiTech & 1e-4     & 12       & 512×512 \\ \hline
\end{tabular}
\label{tab:Training_p}
\vspace{-0.4cm}
\end{table}

\subsection{Evaluation Metrics}

Similarly to other SoTA methods~\cite{li2018csrnet,xu2019autoscale,bai2020adaptive,xiong2019open,varior2019multi,jiang2020attention,thanasutives2021encoder, zhu2019dual}, the mean average error (MAE) and the mean squared error (MSE) are used to evaluate the performance of our architecture. Let $K$ denote the number of test images, $P_k$ the predicted crowd count of the $k$th image, and $G_k$ its ground truth.  Then, $MAE$ and $MSE$ are defined as follows:
\begin{equation}
\footnotesize
 MAE=\frac{1}{K}\sum\limits_{k=1}^K {\lvert P_k - G_k\lvert},
MSE=\sqrt{\frac{1}{K}\sum\limits_{k=1}^K {(P_k - G_k)^2}}.
\end{equation}

\subsection{Performance Comparisons {\em w.r.t.} Loss Functions} 

\begin{figure}[t]
\centerline{
  \includegraphics[width=1.14\linewidth]{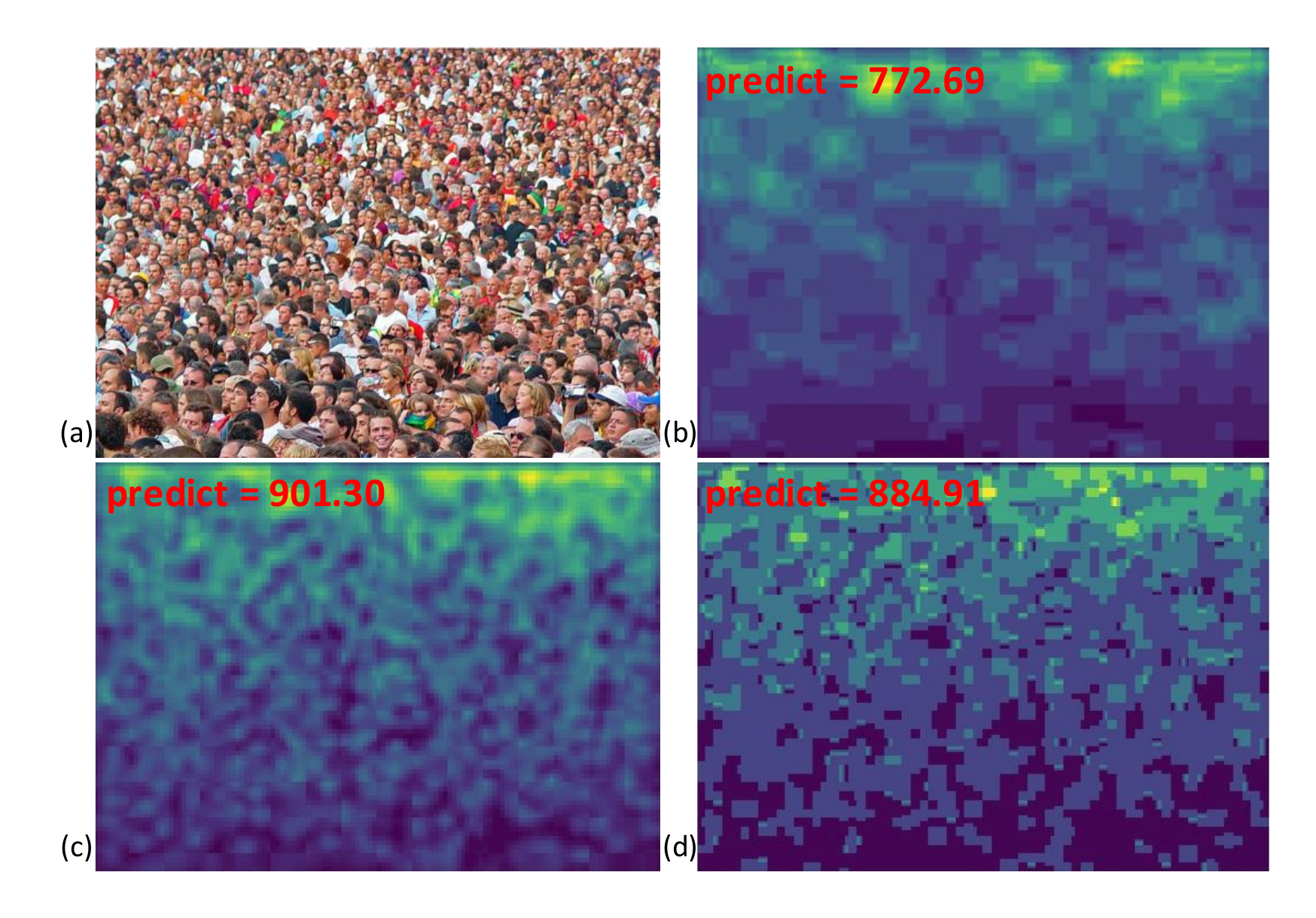}
  \vspace{-0.3cm}
}
\caption{Visualization crowd counting heatmaps generating using different loss functions. (a) Input image (GT = 855, ShanghaiTech Part A). (b) MSE loss. (c) Bayesian loss. (d) Our scale-aware loss.
\vspace{-0.2cm}
}
\label{fig:visualization_compar_loss}
\end{figure}

\begin{figure}[t]
\centerline{
  \includegraphics[width=0.5\textwidth]{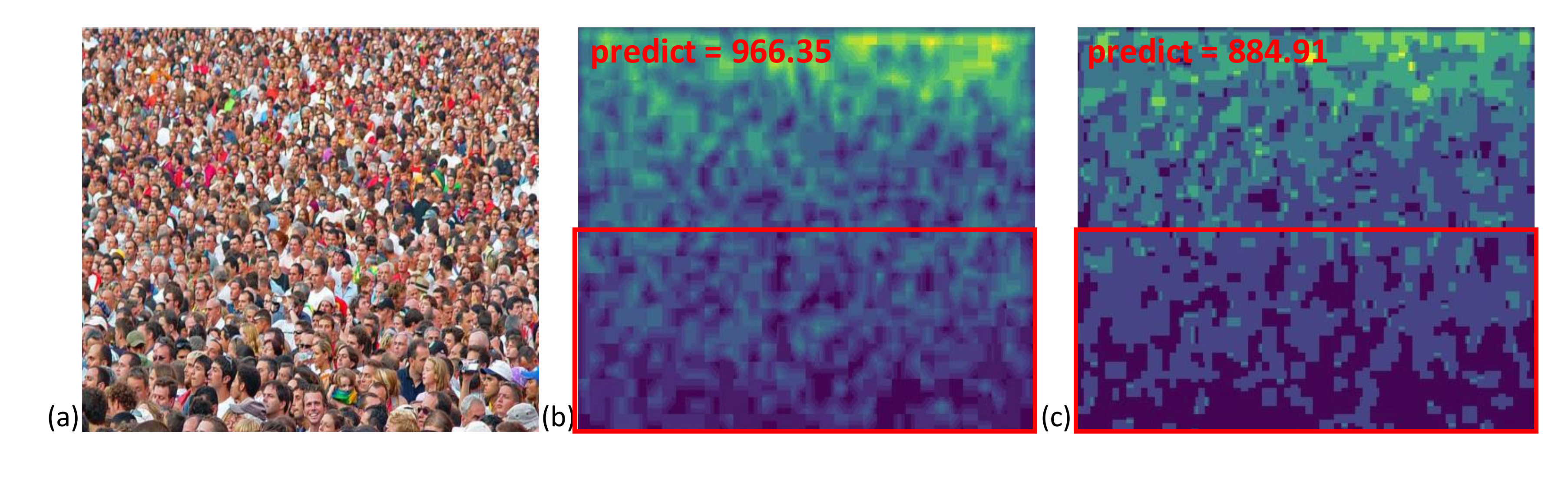}
  
}
\caption{Comparison of heat map visualization between the noiseCC loss and our method. (a) Input image (GT = 855, ShanghaiTech Part A). (b) NoiseCC loss. (c) Our scale-aware loss.
\vspace{-0.2cm}
}
\label{fig:compar nocc noise and our}
\end{figure}


\begin{table}[t]
\caption{Accuracy Comparisons among different Loss Functions
with various Backbones on UCF-QNRF.
\vspace{-0.2cm}
}
\setlength{\tabcolsep}{0.8mm}{
\begin{tabular}{c|cc|cc|cc}
\hline
\multirow{2}{*}{} & \multicolumn{2}{c|}{VGG19} & \multicolumn{2}{c|}{CSRNet} & \multicolumn{2}{c}{MCNN} \\
                  & MAE         & MSE          & MAE          & MSE          & MAE         & MSE        \\ \hline
L2                & 98.7        & 176.1        & 110.6        & 190.1        & 186.4       & 283.6      \\
BL \cite{ma2019bayesian}               & 88.8        & 154.8        & 107.5        & 184.3        & 190.6       & 272.3      \\
NoiseCC\cite{wan2020modeling}          & 85.8        & 150.6        & 96.5         & 163.3        & 177.4       & 259.0      \\
DM-count\cite{wang2020distribution}          & 85.6        & 148.3        & 103.6        & 180.6        & 176.1       & 263.3      \\
Gen-loss\cite{Wan_2021_CVPR} & 84.3        & 147.5        & 92.0         & 165.7        & 142.8       & 227.9      \\ \hline
Ours              & \textbf{83.47}       & \textbf{140.34}       & \textbf{90.83}  & \textbf{150.67}       & \textbf{134.52}      & \textbf{213.71}    
\end{tabular}}
\label{tab:COM_diff loss}
\vspace{-0.4cm}
\end{table}

To evaluate the effectiveness of the proposed loss function, we compare it with L2, which is the most popular loss function, BL  \cite{ma2019bayesian}, NoiseCC \cite{wan2020modeling}, DM-count \cite{wang2020distribution}, and generalized loss \cite{Wan_2021_CVPR} under different backbones.  Table \ref{tab:COM_diff loss} shows the performance comparisons among different loss functions with different backbones. Clearly, our proposed scale-aware loss function outperforms other SoTA loss functions even under different backbones. Since human head sizes are different, the same annotation error causes different effects to degrade the accuracy of crowd counting. Although NoiseCC \cite{wan2020modeling} has pointed out that the noise from the annotation will affect the accuracy of crowd counting, it does not take into account this scaling effect.  Our scale-aware loss function can fix this problem and performs significantly better than other most-adopted loss functions in the UCF-QNRF database.

\subsection{Comparisons with SoTA Methods} 
\begin{table*}[t]
\centering
\caption{Comparison with state-of-the-art crowd counting methods.
\vspace{-0.2cm}
}
\setlength{\tabcolsep}{1.5mm}{
\begin{tabular}{c|c|cc|cc|cc|cc|cc}
\hline
\multirow{2}{*}{Methods}  &\multirow{2}{*}{Venue} &\multicolumn{2}{c|}{UCF-QNRF} & \multicolumn{2}{c|}{NWPU} & \multicolumn{2}{c|}{S. H. Tech-A} & \multicolumn{2}{c|}{S. H. Tech-B} & \multicolumn{2}{c}{UCF CC 50} \\
                             &      & MAE           & MSE           & MAE            & MSE           & MAE                & MSE                 & MAE                & MSE         & MAE                & MSE         \\ \hline
CSRNet\cite{li2018csrnet}      &CVPR'18                       & -             & -             & 121.3          & 522.7         & 68.2               & 115.0               & 10.3               & 16.0         & 266.1             & 397.5        \\
CAN\cite{liu2019context} &CVPR’19  & 107  & 183  & -  & - & 62.3               & 100.0               & 7.8                & 12.2         & 212.2             & 243.7        \\
S-DCNet\cite{xiong2019open}    &ICCV’19                        & 104.4         & 176.1         & -      & -     & 58.3               & 95.0                & 6.7                & 10.7        & 204.2             & 301.3         \\
SANet\cite{cao2018scale}&ECCV’18   & -  & - & 190.6  & 491.4 & 67.0  & 104.5  & 8.4                & 13.6          & 258.4             & 334.9       \\
BL\cite{ma2019bayesian}  &ICCV’19                              & 88.7          & 154.8         & 105.4          & 454.2         & 62.8               & 101.8               & 7.7                & 12.7         & 229.3             & 308.2        \\
SFANet\cite{zhu2019dual}&-  & 100.8          & 174.5         & -          & -         & 59.8               & 99.3                & 6.9                & 10.9       & -             & -          \\
DM-Count\cite{wang2020distribution}&NeurIPS’20                      & 85.6          & 148.3         & 88.4          & 498.0         & 59.7               & 95.7                & 7.4                & 11.8        & 211.0             &  291.5         \\
RPnet\cite{zhang2015cross} &CVPR’15                          & -          & -         & -     & -         & 61.2   & 96.9   & 8.1   & 11.6       & -             & -          \\
AMSNet\cite{Hu2020NAS}&ECCV’20  & 101.8  & 163.2         & -     & -         & 56.7.2   & 93.4   & 6.7   & 10.2     & 208.4             &  297.3            \\
M-SFANet\cite{thanasutives2021encoder}&ICPR'21                          & 85.6          & 151.23        & -         & -             & 59.69              & 95.66              & 6.38               & 10.22         & 162.33             & 276.76       \\
TEDnet\cite{jiang2019crowd}&CVPR’19                             & 113.0         & 188.0         & -          & -         & 64.2               & 109.1               & 8.2                & 12.8        & 249.4             & 354.5         \\
P2PNet\cite{song2021rethinking}  &ICCV’21                            & 85.32         & 154.5         & 77.44          & 362         & 52.74               & 85.06               & 6.25                & 9.9         & 172.72             & 256.18        \\
	
GauNet\cite{cheng2022rethinking} &CVPR’22                            & 81.6         & 153.7          & -          & -         & 54.8               & 89.1               & 6.2                & 9.9          &  186.3             & 256.5       \\

\hline

SFP-Net(BL Loss) &-                          & 85.42         & 145.44        & 86.72         & 442.9        & 55.28              & 90.37               & 6.5               & 10.68         & 167.48             & 235.41       \\ 
SFP-Net(our loss) &-                             &\textbf{78.36}         & \textbf{124.25}         & \textbf{75.52}          & \textbf{349.73}         & \textbf{52.19}               & \textbf{76.63}               & \textbf{6.16}                  & \textbf{9.71}  & \textbf{150.66}             & \textbf{187.89}
\\ 
 \hline
\end{tabular}}
\label{tab:COM_ALL}
\vspace{-0.4cm}
\end{table*}
To further evaluate the performance of our proposed method, eleven SoTA methods are compared here for performance evaluation; that is,
CSRNet\cite{li2018csrnet}, CAN\cite{liu2019context}, S-DCNet\cite{xiong2019open}, SANet\cite{cao2018scale}, BL\cite{ma2019bayesian}, SFANet\cite{zhu2019dual}, DM-Count\cite{wang2020distribution}, RPnet\cite{zhang2015cross}, AMSNet\cite{Hu2020NAS} M-SFANet\cite{thanasutives2021encoder}, TEDnet\cite{jiang2019crowd}, P2PNet\cite{song2021rethinking}, and GauNet\cite{cheng2022rethinking}. Table \ref{tab:COM_ALL} shows 
the comparative results among these SoTA methods in the four benchmark data sets above. Clearly, our method achieves
significantly better performance, especially for large-scale datasets such as UCF-QNRF, NWPU-Crowd, and ShanghaiTech Part A. In addition, our model achieves the best MAE and also the best MSE on all the above datasets.  However, our method outperforms GauNet not signifcantly on ShanghaiTech Part B.  The limitation of our method requires more training samples to train our model.  When the dataset is small, our method will overestimate the covariance matrix of annotation errors. 
\subsection{Ablation Studies}
\begin{figure}[t]
\centerline{
\includegraphics[width=1.1\linewidth]{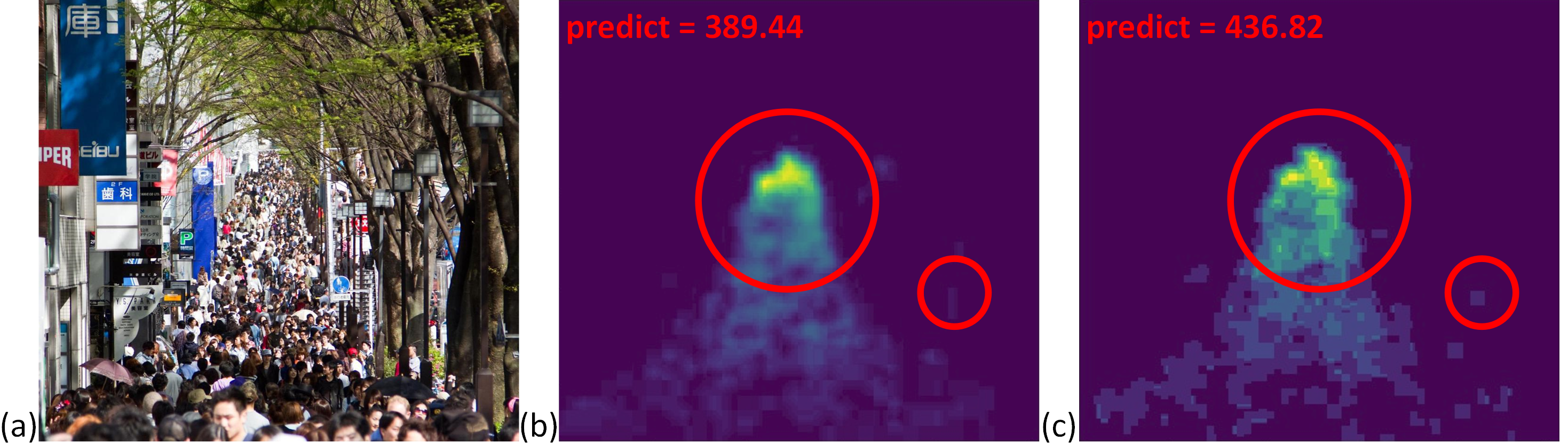}
\vspace{-0.3cm}
}
\caption{Visualization of the SFP-Net (b) without and (c) with intra-fusion, where (a) shows the input image with GT = 429 from ShanghaiTech Part A.}
\label{fig:intra}
\vspace{-0.6cm}
\end{figure}

\begin{table}[t]
\caption{Ablation study of Intra fusion ,and differnt Scale demsity on UCF-QNRF dataset.
\vspace{-0.2cm}
}
\setlength{\tabcolsep}{0.5mm}{
\begin{tabular}{c|c|ccc|cc}
\hline
\multirow{2}{*}{Methods}      & \multirow{2}{*}{SFM+IFM} & \multirow{2}{*}{Scale1} & \multirow{2}{*}{Scale2} & \multirow{2}{*}{Scale3} & \multicolumn{2}{c}{UCF-QNRF} \\
                              &                               &                          &                          &                          & MAE        & MSE         \\ \hline
\multirow{6}{*}{SFP-Net} & \multirow{3}{*}{}             & \checkmark                        &                          &                          & 85.45      & 145.74      \\
                              &                               & \checkmark                        & \checkmark                        &                         & 84.07      & 135.63      \\
                              &                               & \checkmark                        & \checkmark                        & \checkmark                       & 82.42      & 130.04      \\ \cline{2-7} 
                              & \multirow{3}{*}{\checkmark}            & \checkmark                        &                          &                         & 83.81      & 140.19      \\
                              &                               & \checkmark                        & \checkmark                        &                         & 82.71      & 130.29      \\
                              &                               & \checkmark                        & \checkmark                        & \checkmark                       & \textbf{78.36}      & \textbf{124.25}      \\ \hline
\end{tabular}}
\label{tab:ablation}
\vspace{-0.4cm}
\end{table}

 To demonstrate the effectiveness of our fusion approach, we conducted an ablation study on how the addition of ``fusion'' and the number of scales used improve the accuracy of crowd counting. Most of the objects in the UCF-QNRF dataset are smaller than those in other dataset. Thus, UCF-QNRF is adopted here to fairly evaluate the effect of our proposed fusion module.
In Table 
\ref{tab:ablation}, we can see that using our fusion module is significantly better than not using it.
For example, our SFP-NET with this module reduces the error rates significantly from 82.42 to 78.36 in MAE and from 130.04 to 124.25 in MSE for the UCF-QNRF dataset.

We also evaluated the effects of the number of scales on improving the accuracy of crowd count.  There are five pooling layers created in VGG19 that cause the original image to be scaled down to only 1/32$\times$1/32 ratio.  The feature map at the last one layer cannot provide enough information to calculate the required covariance matrix. The first layer is too primitive for crowd counting. Thus, using 3 layers provides optimal performance results. So in Eq.(\ref{eq:all-density}) we setting $S$ = 3 .  Table 
\ref{tab:ablation} shows the accuracy comparisons between three combinations of three scales (corresponding to layer 2, layer 3, layer 4).  The three-scale scale-aware loss function significantly improves the accuracy of crowd counting in the UCF-QNRF dataset, especially in the MAE metric.

\begin{figure}[t]
\centerline{
  \includegraphics[width=0.8\linewidth]{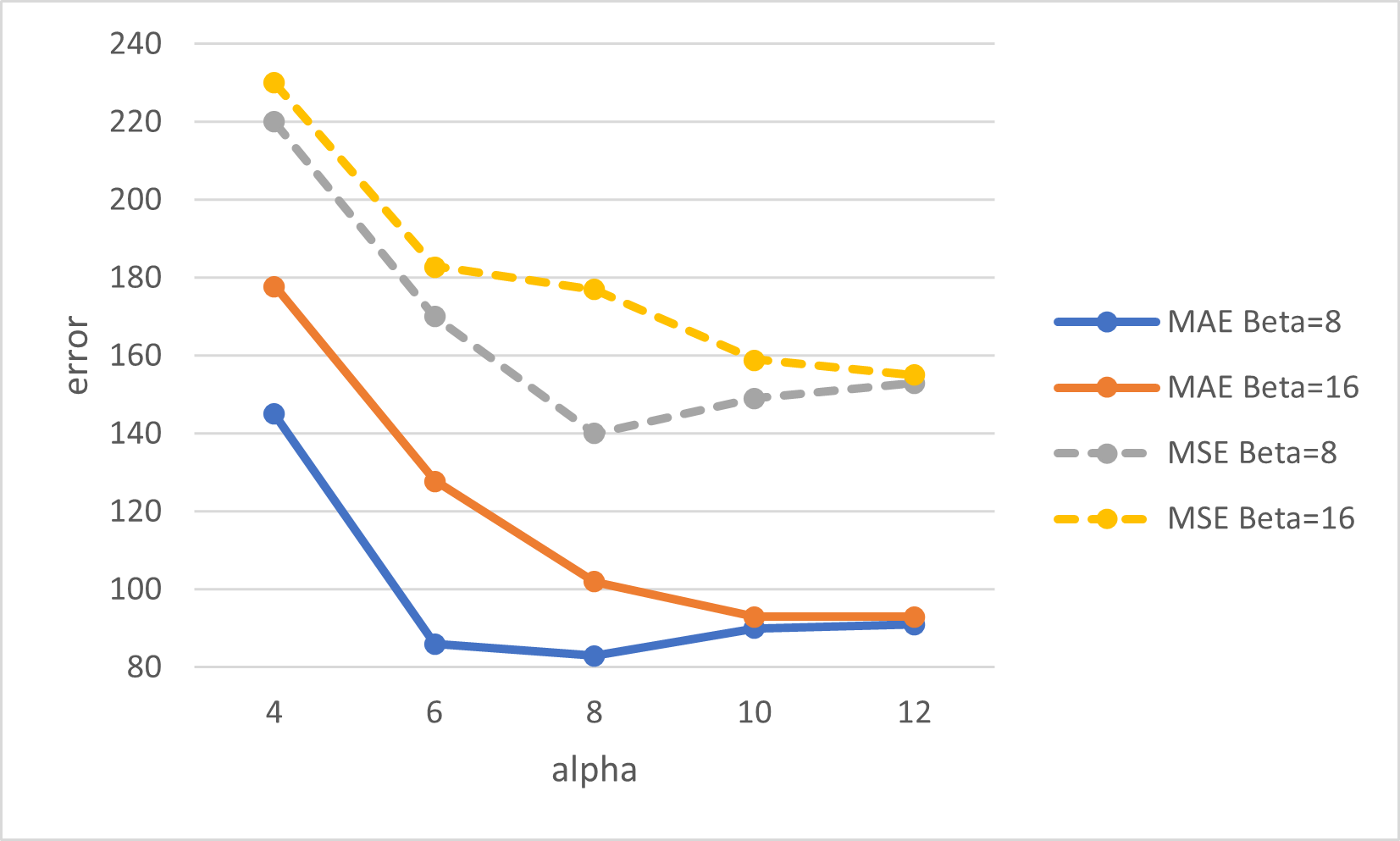}}
\caption{
 MAE changes when different annotation variance $\alpha$ and density map variance $\beta_{1}$ were set. Clearly, when $\beta_{1}=8$  and $\alpha=8$, the MAE is the lowest.
\vspace{-0.4cm}
}
\label{fig:effect alpha}
\end{figure}

{\bf Effects of variance parameters $\alpha$ and $\beta_s$:} We also conducted an experiment to investigate the effect of variance in annotation $\alpha$ and variance in the density map $\beta$. As shown in Figure~\ref{fig:effect alpha}, when $\beta$ increases, the MAE decreases, but when $\beta\ge 8$, the MAE starts to increase instead, so we set $\beta_1=8$. Then, $\beta_{s+1}=\beta_s/2$. Regarding $\alpha$, when $\alpha$ is very small, the MAE is large, and when $\alpha\ge8$, the MAE decreases and tends to be stable, so we set $\alpha=8$.

{\bf Visualization results of heat maps.}  Figure~\ref{fig:visualization_compar_loss} shows the visulization results when different loss functions were used. The ground truth of heads in (a) is 855.  The heat maps generated by the MSE loss and the Bayesian loss~\cite{ma2019bayesian} are visualized in (b) and (c) with the predicted results 772.69 and 901.30, respectively.  Clearly, the MSE loss performs better than the BL method. (d) is the visualization result generated by our scale-aware loss with the predicted value 884.1.  Compared to (b) and (c), our loss function can generate a more detailed heat map for heads, since the annotation errors are taken into account in crowd modeling.  Thus, our method outperforms the above two methods.  NoiseCC~\cite{wan2020modeling} also points out that noise from annotation will affect the accuracy of crowd counting.  Figure~\ref{fig:compar nocc noise and our}(a) and (b) show the visualization results of heat map generated by NoiseCC and our loss function with the predicted head numbers 966.35 and 884.91, respectively. Since the scaling effect of annotation errors is not considered in NoiseCC, the heat map generated by NoiseCC is more blurred than that generated by our method. Thus, a better accuracy is obtained by our scale-aware method.   Figure~\ref{fig:intra} shows the visualization results generated by our method without/with the fusion module.  (b) and (c) are the visualization results without/with the fusion module, respectively. The fusion model generates various synthetic layers so that better density maps can be constructed for crowd counting.  Clearly, in (c), a more detailed heat map for small heads was generated, which leads to a better crowd-counting accuracy than in (b).                           

\section{Conclusion}

In this paper, we propose a novel loss function for different-scale annotation noise modeling, which has been proven to more accurately predict the number of people in a crowd. The loss function can be decomposed into one that considers pixel correlations at different scales, which models the correlation between pixels, and a regression term that ensures that all densities sum to 1. First, we assume that the annotation noise follows a Gaussian distribution. Then, derive the probability density function for the density values. We use the full covariance Gaussian and low-rank approximation to approximate the joint distribution of density values to reduce the computational cost and discuss the problems of different scales and use different standard deviations to approximate the annotation errors at the front, middle, and back areas. For people with larger foreground scales, ''intra-layer'' fusion has also been shown to have a significant effect on small object discrimination in the UCF-QNRF dataset. Finally, our proposed SFP-Net outperforms the SoTA methods in all datasets. Our MAE=78.36 on the UCF-QNRF dataset is currently the only method below 80. The effectiveness of the proposed loss function is confirmed. Under the same model, compared to several existing mainstream loss functions, our proposed loss function outperforms them.

{\bf Future Work.} This paper sets the parameters $\alpha$ and $\beta_{s}$ by humans.  Clearly, they can be automatically learned from the data.  In the near future, they will be automatically learned to construct better heat maps for crowd counting.  Another future work is to propose a lightweight SFP-Net for working directly on drones.


{\small
\bibliographystyle{ieee_fullname}
\bibliography{egbib}
}

\end{document}